\documentclass{llncs}
\usepackage{llncsdoc}
\usepackage{amsmath}
\usepackage{longtable}
\usepackage{array}
\begin{document}
\markboth{Identifying Risk Factors for Prescription Drug Side Effects
}{Identifying Risk Factors for Prescription Drug Side Effects}
\thispagestyle{empty}

\title{Identifying Candidate Risk Factors for Prescription Drug Side Effects using Causal Contrast Set Mining}


\author{
Jenna Reps,
Zhaoyang Guo,
Haoyue Zhu,
Uwe Aickelin
\email{jenna.reps@nottingham.ac.uk} }
\institute{School of Computer Science \\
University of Nottingham\\
Nottingham, NG8 1BB}


%

\maketitle

\begin{abstract}
Big longitudinal observational databases present the opportunity to extract new knowledge in a cost effective manner. Unfortunately, the ability of these databases to be used for causal inference is limited due to the passive way in which the data are collected resulting in various forms of bias. In this paper we investigate a method that can overcome these limitations and determine causal contrast set rules efficiently from big data. In particular, we present a new methodology for the purpose of identifying risk factors that increase a patient’s likelihood of experiencing the known rare side effect of renal failure after ingesting aminosalicylates. The results show that the methodology was able to identify previously researched risk factors such as being prescribed diuretics and highlighted that patients with a higher than average risk of renal failure may be even more susceptible to experiencing it as a side effect after ingesting aminosalicylates.

\end{abstract}

\section{Introduction}
Longitudinal observational data potentially hold a wealth of information, however we are currently limited in the ability to efficiently extract causal relationships from this form of data due to bias and confounding \cite{giordano2008limits}. In randomised clinical trials confounding can be overcome by manipulating the variables and mixing the potential confounders equally between the group given the drug and the control group. Unfortunately, this is not possible for observational data as the data are passively observed. As a consequence, spurious results are common when analysing observational data due to the various forms of bias in the data. In the medical field the gold standard for causal discovery are randomised clinical trials \cite{cochran1973controlling}. However, these are costly and sometimes unethical \cite{black1996we}. If medical longitudinal observational data could be successfully analysed and the results used to complement randomised trials for causal discovery, then this would address these issues. This would enable a greater understanding of various medical mechanisms and enhance current knowledge. 

Bayesian causal discovery techniques that learn complete causal models have often been used to identify causal relationships in longitudinal observational data\cite{cooper1992bayesian}. Due to scalability issues the recent focus has shifted towards constraint based methods \cite{silverstein2000scalable}. Although the constraint based methods have performed well in some domains, they rely on numerous assumptions \cite{heckerman1999bayesian} that may not always hold true and may still be inefficient for data with high volume and high variety. A recent approach for identifying causal association rules included a two step method, of firstly mining association rules and secondly implemented a cohort study to filter out those that are likely to be causal. This was accomplished by identifying controls that had the antecedent and matched specific attributes of the cases. The odds ratio was then used as the filter, as only the rules with a significant deviation between how often the consequence occurred for the cases and controls were kept \cite{li2013mining}. In this paper we attempt a similar approach for identifying causal contrast sets but use logistic regression as a filter. Rather than using the odds ratio, we use the p-values of the logistic regression variables to indicate how significant having the antecedent is for the occurrence of the consequence. As the logistic regression can consider covariates such as age, and gender into the model, we can filter contrast set rules that are caused by observed confounders. 

In this paper we present a proof-of-concept candidate risk factor detection algorithm based on causal contrast set mining. Causal contrast set mining is a term we use to define the discovery of causal association rules that identify differences between various groups. The algorithm firstly identifies interesting rules consisting of sets of events that commonly precede a user specified event and then investigates how often these interesting rules occur in general. Rules that occur more often before the user specified event are then investigated via a logistic regression model. This reduces age/gender confounding and highlights the most interesting rules. We implement the methodology to a real word dataset. The dataset we use is a UK general practice database containing complete medical and drug prescription records for millions of patients within the UK. Our focus is towards identifying risk factors for patients' experiencing prescription drug side effects for the drug family aminosalicylates (5-ASAs). These drugs are often given to treat inflammatory bowel disease but are known to cause renal failure with an incidence rate of 0.17 cases per 100 patients per year \cite{van20045}. The purpose of this research is to investigate a new technique for mining contrast set causal relationships efficiently and evaluate its potential for identifying candidate risk factors of patients experiencing side effects to prescribed medication.

\section{Materials \& Methods}
\subsection{The Health Improvement Network}
The Health Improvement Network (THIN) database (www.thin-uk.com) is a large longitudinal observational database containing medical records for millions of patients within the UK. There are over 600 general practices within the UK that are registered to the scheme consisting of over 3.5 million active patients. For each patient within THIN, their demographics such as age, gender and location are known, as well as their complete medical and therapy record histories during the period of time they are registered at a participating practice. The suitability of this database for epidemiological study has been investigated and the results show it is reasonably representative of the general UK population \cite{THIN}. It is worth highlighting that the database does have some potential issues, such as not containing over the counter prescriptions, only containing data that patients have told their doctors about and delays in the recording of medical event into the database. A common problem with the database is historical event dropping, when a patient moves general practices, it is common for the patient to have historical illnesses/events recorded shortly after registering. To prevent this biasing analyses, it is standard to exclude the first year of a patient's records after moving to a new general practice \cite{Lewis2006}. This preprocessing was implemented in this study.

The READ code system is the coding system used within UK primary care to record medical events \cite{READ}. Each READ code corresponds to a medical event (e.g., a diagnosis, an administrative event, a laboratory result or a symptom). The READ codes consist of 5 alphanumeric digits and have a hierarchal tree structure based on the level of detail of the corresponding medical event being recorded. The level of a READ code corresponds to how many non dot digits it contains, for example the READ code `A10..' is a level 3 READ code, whereas the READ code `A....' is a level 1 READ code. A level 2 READ code is the child of a level 1 READ code if the READ codes have the same first digit. This is generalised to a level $n\in \{2,3,4,5\}$ READ code being the child of the level $n-1$ READ code if the first $n-1$ digits of both READ codes are the same. The advantage of this hierarchal structure is that a child READ code represents a more specific version of its parent READ code's corresponding medical event. For example, the READ code `A....' corresponds to the description `Infection' and is the parent of the READ code `A1...' corresponding to `Tuberculosis', which is the parent of the READ code `A11..' corresponding to `Pulmonary tuberculosis'.

Prescriptions are recorded into THIN using a drug code and each prescription also contains the drug's British National Formula (BNF) code \cite{BNF}. The BNF code groups drugs into similar families. Each prescription can be linked to up to three BNF codes. 

\subsection{Algorithms}
\subsubsection{Association rules mining}
Association rules mining \cite{Agrawal1993} is a method for discovering relations between variables in large databases. It was originally designed to identify relationships between items that are commonly purchased together (occur in the same shopping baskets). The relations are normally of the form \{antecedent events \} $\to$ \{consequence\}, meaning that if we find all of the antecedent events in a shopping basket, then we have a good chance of finding the consequence. An example of an association rule is \{milk, butter\} $\to$ \{bread\}, which means shoppers that buy milk and butter are also likely to buy bread. 

The search space for identifying association rules can be extremely large with big datasets. Therefore it is common to restrict the search to only include rules containing sets of items that appear frequently in baskets. This is accomplished by specifying a minimum support threshold, and only items/itemsets that occur more often than the support are considered. These are referred to as frequent itemsets.

Formally, let $I = \{i_{1}, i_{2},..., i_{n}\}$ be a set of $n$ items and $t = X \subset I$ be a transaction containing a set of items. We denote the database by $D = \{ t_{1}, t_{2},..., t_{m}\}$. This is a set of m transactions. The support of an itemset $X$ is the proportion of transactions within the database that contain X,
\begin{equation}
supp(X) = |\{ t_{i} \in D | X \subset t_{i} \} |/m
\end{equation}
An itemset $X$ is said to be frequent if its support is greater than a given threshold $supp(X)>\omega$, where $\omega$ is called the minimum support. 

The confidence of an association rule $X \to Y$ is the fraction of baskets that contain both $X$ and $Y$ ($supp(X \cup Y)$) divided by the number of baskets containing $X$ ($supp(X)$),
\begin{equation}
conf(X \to Y) = supp(X \cup Y)/supp(X)
\end{equation}
this is similar to the conditional probability of $Y$ given $X$. In general, the association rules $X \to Y$ are identified such that the support and confidence of $X \to Y$ are greater than the minimum support and confidence thresholds.

There are various methods for identifying contrast set rules, including discovering emergent patterns by considering the ratio of two supports \cite{dong1999efficient}, using a suitable search technique combined with statistical hypothesis testing \cite{bay2001detecting} or creatively using a classifier \cite{novak2009supervised}. Emergent pattern discovery is suitable for simple problems that only require contrasting two groups. This is what we will do to identify candidate risk factors, as we just need to compare the patients that experienced the adverse drug reaction with those that did not.

\subsubsection{Logistic Regression}
Logistic regression \cite{hosmer2004applied} is a method that expresses the log odds of belonging to a class as a linear combination of the features,
\begin{equation}
ln(P(Y|\mathbf{X})/(1-P(Y|\mathbf{X}))) = w_{0} + \sum_{i} w_{i}X_{i} 
\end{equation}
The parameters $w_{i}$ are found using maximum likelihood. This is re-arranged to give the conditional probability of belonging to each class as,
\begin{equation}
\begin{split}
P(Y=0|\mathbf{X}) &= \frac{exp(w_{0} + \sum_{i} w_{i}X_{i} )}{1+exp(w_{0} + \sum_{i} w_{i}X_{i} )} \\
P(Y=1|\mathbf{X}) &= \frac{1}{1+exp(w_{0} + \sum_{i} w_{i}X_{i} )} \\
\end{split}
\end{equation}
therefore, class $0$ is chosen when $exp(w_{0} + \sum_{i} w_{i}X_{i} ) >1$ and 1 is chosen otherwise. The parameter $w_{i}$ and its standard error of the logistic regression tell us how significant the i\textsuperscript{th} feature, $X_{i}$, is in determining the class. In this paper we use a significance level of 5\%.

\subsection{Methodology}
The proposed candidate risk factor identification methodology consists of four steps. The first step is creating two different databases based on whether a patient who was prescribed a 5-ASA experienced renal failure or not. The second step is to identify frequent itemsets for the patients who experience renal failure after 5-ASAs and calculate whether these itemsets occur more often for these patients than for the patients prescribed 5-ASAs in general. This identifies any potential risk factors that are common (occur in more than 5\% of the patients). The third step is to identify whether these potential risk factors are a significant influence on experiencing renal failure after a 5-ASA when accounting for age and gender confounding. The final step is presenting the frequent itemsets that occur more than in general for the patients who experience renal failure after a 5-ASA ordered by the p-value indicating the significance of the itemset's presence in predicting the chance of renal failure after a 5-ASA.

\subsubsection{Step 1: Partition Databases}
Similar to market baskets, patients’ medical baskets can be constructed based on the records they have in the THIN database and frequent itemset mining can be applied to find frequent medical events sets. Due to the number of possible itemsets being very large, frequent itemset mining is often restricted so that only interesting itemsets are discovered.

To generate association rules for the THIN database we consider the items to be all the medical events and all the drugs recorded within the THIN database. So the THIN items are $I = \{$all the medical events and all the drugs\} and a transaction is $X \subset I$. Then we generated two databases from the THIN database: $D1$ contains the itemsets of patients that took 5-ASA but did not suffer from renal failure within a month and $D2$ contains the itemsets of patients that took 5-ASA and suffered from renal failure within a month. For each transaction, $t_{i}^{D1} \in D1$ or $t_{i}^{D2} \in D2$, the transaction consists of all the items within the THIN database that are recorded for the i\textsuperscript{th} patient in the database.

For example, if a patient had renal failure recorded within a month of a 5-ASA and only had the READ codes 681.., 8CB.., 9R8.., 246.. and H33..00 recorded in THIN, then his corresponding transaction in $D2$ would be $\{$681..,8CB..,9R8.., 246.., H33..$\}$. 

\subsubsection{Step 2: Calculating Support Ratio}
In general the THIN data is sparse and the majority of items have a low support. However, to identify risk factors for renal failure after ingesting a 5-ASA we only need to investigate the itemsets that are frequent in the patients that took 5-ASA and suffered from renal failure (frequent itemsets in $D2$). Then we need to find which of these frequent itemsets from $D2$ have a higher support than within $D1$, as this indicates itemsets that are more common in the 5-ASA patients who experience renal failure compared to all the 5-ASA patients. Therefore, we apply frequent itemset mining to the database $D2$ with minimum supports of $\omega=0.05$ and for each frequent item we also calculated its support in $D1$. We then calculate the support ratio for each frequent itemset $X$ from $D2$,
\begin{equation}
suppRatio(X) = [|\{ t_{i} \in D2 | X \subset t_{i} \} |/m_{2}]/ [|\{ t_{i} \in D1 | X \subset t_{i} \} |/m_{1}]
\end{equation}
where $m_{1}$ and $m_{2}$ are the number of patients that took 5-ASA but did not suffer from renal failure and took 5-ASA and suffered from renal failure, respectively. The value $\omega=0.05$ was chosen as this means that any identified risk factors occur for at least 5\% of the patients experiencing renal failure after 5-ASA. Therefore we are identifying common risk factors, however this value can be adjusted.

After applying the association rules, we will get a table containing the frequent itemsets of $D2$ and their support in both D1 and D2. The rate of each frequent itemset corresponds to the ratio of two support values (support(X,ASA$\to$RF) / support(X,ASA$\to \neg$RF)), see Table \ref{supp}.
\begin{table} \centering
\caption{Example of how to calculate the suppRatio for each frequent itemset. }
\label{supp}
\begin{tabular}{cccc}
Itemset (X) & Support(X,ASA$\to$RF) & Support(X,ASA$\to\neg$RF) & suppRatio(X) \\ \hline \hline
$\{$G2...$\}$ & 0.15903 & 0.056378 & 2.820757 \\
$\{$G3...$\}$ & 0.080863 & 0.028041 & 2.883717 \\
$\{$6781.,G2...00$\}$ & 0.067385 & 0.023302 & 2.891863 \\
$\{$D21z.$\}$ & 0.067385 & 0.029588 & 2.277463 \\
$\{$65E..$\}$ & 0.078167 & 0.036105 & 2.165022 \\
... & & & \\
\end{tabular}
\end{table}
The itemsets with a suppRatio greater than 1 are considered potential risk factors that will be further evaluated using logistic regression.

\subsubsection{Step 3: Logistic Regression}
We then applied logistic regression with the independent variables: presence of potential risk factor, presence of 5-ASA, age and gender and dependant variable indicating renal failure. This identified whether the potential risk factors are in fact significant risk factors for experiencing renal failure after 5-ASAs when accounting for age/gender confounding.

To apply the logistic regression we needed to consider a set of cases (the patience with renal failure recorded in THIN) and a set of controls (the patients with no renal failure recorded in THIN). For each patient experiencing renal failure we selected 5 controls who did not.  Increasing the number of controls per case is a technique that can increase the power of the analysis and 5 controls per case were chosen as we have a large number of controls available but only a limited number of cases. For each case, the age used in the logistic regression is considered as the age when the case first suffered from renal failure in life. Each control was selected by picking a random non-renal failure patient and a random point in the time while the patient is active in THIN such that the age/gender distributions of the cases and controls were the same. 

Then, for each potential risk factor frequent itemset identified in step 2 (each $X$) we created the case/control data as displayed in Table \ref{lg},
\begin{table} \centering
\caption{Example of the data used for each logistic regression.}
\label{lg}
\begin{tabular}{cccccc}
PatientId & Age & Gender & $X$ & ASA & RF \\ \hline \hline
1 & 45 & 1 & True & True & True \\
2 & 50 & 2 & False & True & False \\
3& 45 & 1 & False & True & True \\
4& 59 & 2 & False & True & False \\
5 & 22 & 2 & True & False & True \\
... & & & & &\\
\end{tabular}
\end{table} 
where the variable $X$ is True if the patient's itemset up to their specified age contains $X$, the variable ASA is True if the patient was prescribed a 5-ASA before the specified age and RF is True if the patient has renal failure recorded in THIN and False otherwise. The logistic regression with RF as the dependant variable was then applied considering the independent variables: age, gender, $X$, and ASA. The interaction between the ASA variable and the $X$ variable was also included. 

\subsubsection{Step 4: Ranking}
The p-value of the interaction between the frequent itemset and 5-ASA was calculated to evaluate whether the frequent itemset is a risk factor of experiencing renal failure after 5-ASA. The smaller the p-value is, the greater the confidence that the frequent itemset corresponds to a risk factor. The p-value of each frequent itemset is extracted and listed in the result table. The results are returned ordered by the p-values in ascending order. 
\begin{table} \centering 
\caption{Example of the output of the methodology.}
\label{output}
\begin{tabular}{cccc}
Itemset (X) & P-value(Age) & P-value(Gender) & P-value(ASA*Rules) \\ \hline \hline
$\{$9N1O.$\}$ & 8.25E-8 & 3.08E-1 & 2.78E-18 \\
$\{$G33..$\}$ & 1.87E-8 & 2.06E-1 & 2.28E-44 \\
... & & &\\
\end{tabular}
\end{table}
The final output of the methodology is this ranked list of frequent itemsets as illustrated in Table \ref{output}.

\subsection{Software}
We use SQL to manage the data and R \cite{R} to perform the analysis. The package arules \cite{arules} was used to identify the frequent itemsets.
\newpage

\section{Results \& Discussion}
\begin{longtable}{>{\raggedright}p{2.8cm}ccccp{2.3cm}}
\caption{The results of the candidate risk factor identification for the occurrence of renal failure after 5-ASA.}
\label{rf} \\
Description & RFsupp& noRFsupp & suppRatio & p-value & Potential Link \\ 
& (val $\times 10^{-2} $)& (val $\times 10^{-2}$) & & & \\ \hline \hline \endfirsthead
Hypertensive disease & $15.9 $ & $5.64$ & $2.82$ & $1.62 \times 10^{-30}$ & Hypertension\\
Furosemide tabs & $11.9$ & $3.21 $ & $3.70$ & $7.86 \times 10^{-30}$ & \textbf{Diuretics} \cite{de20055} \\
BP reading & $8.63 $ & $2.16$ & $3.99$ & $1.69 \times 10^{-24}$ & Hypertension\\
Co-proxamol tabs & $28.3 $ & $17.4 $ & $1.63$ & $1.16 \times 10^{-23}$ & Pain \\
Rheumatoid arthritis & $24.5 $ & $14.1 $ & $1.74$ & $1.3 \times 10^{-23}$ & Arthritis\\
Blood pressure reading & $9.70$ & $2.92$ & $3.32$ & $1.42 \times 10^{-23}$ & Hypertension \\
Furosemide \& Co-proxamol tabs & $6.74$ & $1.38 $ & $4.89$ & $3.07 \times 10^{-23}$ & Diuretics \& Pain \\
Diabetes mellitus & $8.36 $ & $2.14$ & $3.91$ & $8.22 \times 10^{-23}$ & Diabetes \\
Influenza inactivated split virion vaccine & $9.43$ & $2.70 $ & $3.50$ & $1.1 \times 10^{-22}$ & Influenza vaccination \\
Co-proxamol tabs \& Hypertensive disease & $7.01$ & $1.82 $ & $3.84$ & $4.62 \times 10^{-21}$ & Pain \& Hypertension \\
Pain & $11.9$ & $5.03 $ & $2.36$ & $2.31 \times 10^{-18}$ & Pain \\
Osteoarthritis & $11.1 $ & $4.65$ & $2.38$ & $1.1 \times 10^{-17}$ & Arthritis \\
Co-proxamol tabs \& Pain & $7.82$ & $2.70 $ & $2.89$ & $4.41 \times 10^{-16}$ & Pain \\
Ischaemic heart disease & $8.09$ & $2.80 $ & $2.88$ & $4.51 \times 10^{-16}$ & Hypertension \\
Co-proxamol tabs \& Rheumatoid arthritis & $10.2$ & $4.48$ & $2.29$ & $2.31 \times 10^{-15}$ & Pain \& Arthritis \\
Health education offered \& Hypertensive disease & $6.74$ & $2.33 $ & $2.89$ & $2.45 \times 10^{-14}$ & Hypertension \\
Influenza inactivated surface antigen vaccine & $9.97$ & $4.52 $ & $2.21$ & $5 \times 10^{-14}$ & Influenza vaccination \\
Atenolol tabs & $10.2 $ & $4.68 $ & $2.19$ & $5.4 \times 10^{-14}$ & Hypertension \\
Screening-health check & $9.16 $ & $4.17 $ & $2.20$ & $1.66 \times 10^{-13}$ & \\
Amoxicillin caps \& Hypertensive disease & $6.20 $ & $2.21 $ & $2.80$ & $4.02 \times 10^{-13}$ & Antibiotic \& Hypertension \\
Essential hypertension & $12.9 $ & $7.58 $ & $1.71$ & $2.59 \times 10^{-12}$ & Hypertension \\
Pain \& Screening-general & $6.47$ & $2.47$ & $2.62$ & $5.03 \times 10^{-12}$ & Pain \\
Influenza vaccination & $7.82 $ & $3.61$ & $2.17$ & $5.95 \times 10^{-12}$ & Influenza vaccination \\
Arthritis & $11.1$ & $6.12 $ & $1.81$ & $2.68 \times 10^{-11}$ & Arthritis \\
Anaemia unspecified & $6.74 $ & $2.96 $ & $2.28$ & $5.94 \times 10^{-11}$ & Anaemia \\
Loperamide caps & $7.28 $ & $3.42$ & $2.13$ & $1.27 \times 10^{-10}$ & \textbf{Dehydration} \cite{de20055} \\
Cardiac disease monitoring & $7.01 $ & $3.11 $ & $2.25$ & $1.79 \times 10^{-10}$ & Hypertension \\
Amoxicillin caps \& Pain & $6.20$ & $2.63 $ & $2.36$ & $1.85 \times 10^{-10}$ & Antibiotic \& Pain \\
Paracetamol tabs& $15.4 $ & $10.5 $ & $1.46$ & $2.33 \times 10^{-10}$ & Pain \\
Screening-general \& Rheumatoid arthritis & $7.28$ & $3.46 $ & $2.10$ & $4 \times 10^{-10}$ & Arthritis \\
\end{longtable}

The top 30 antecedents that occur significantly more often for patients who experience renal failure after ingesting a 5-ASA, ordered by the logistic regression p-value, are presented in Table \ref{rf}. The results suggest that some potential risk factors for experiencing renal failure after ingesting a 5-ASA are hypertension, diuretics, pain, arthritis, diabetes, influenza vaccination, anaemia, dehydration and antibiotics. 

The results identified some known risk factors. However, in general there is little information about the risk factors making the evaluation difficult. This highlights the importance of a new methodology for discovering risk factors. In a previous study it was observed that diuretics and dehydration may be risk factors \cite{de20055}. The diuretic drug furosemide was ranked second by the methodology and patients with a history of furosemide were 3.7 times more likely to experience renal failure after 5-ASAs. We found that those with a history of co-proxamol and furosemide were 4.89 times more likely to experience renal failure after 5-ASAs. The drug loperamide was also identified as a risk factor by the method. This drug is used to treat diarrhoea and may indicate that the patients who experienced renal failure after loperamide and 5-ASAs were dehydrated.

Hypertension is a general risk factor for developing renal failure. Interestingly, this research suggests that 5-ASAs increase hypertension suffering patients' susceptibility to renal failure. Therefore 5-ASA may need to be prescribed more carefully to patients who are already susceptible to renal failure. It is common for side effects to occur in patients that have a higher background risk of the event, so this is not unexpected.

Some painkillers and drugs used to treat hypertension are known to cause renal failure. The identification of pain and hypertension as risk factors may indicate an interaction between these drugs and the 5-ASAs that results in the side effect of renal failure. Therefore the methodology may highlight indirect risk factors. This does highlight one limitation of this methodology, it is difficult to identify whether the medical event or the drugs used to treat the medical event may be risk factors. Additional work will be required to determine whether the identified potential risk factor is a direct or indirect risk factor.

It is worth highlighting that this methodology cannot definitively determine the risk factors of known adverse drug reactions. Any results obtained need to be validated via formal epidemiological studies. However, this method can highlight the most likely risk factors and can be considered to be a filter. Therefor this methodology may lead to more efficient discovery of unknown risk factors by identifying which candidate risk factors should be investigated further. Effectively this methodology is an ADR risk factor filter.

In this paper we chose to use a minimum support of $0.05$ as this ensured any identified risk factors occurred for more than 5\% of the patients who experienced the side effect. This value may need to be adjusted based on the type of risk factors of interest or based on how common the side effect being investigated is.

\section{Conclusions}
In this paper we have presented a proof-of-concept of a novel methodology for identifying causal contrast set rules in big longitudinal observational data. The methodology was able to identify known risk factors for patients experiencing renal failure after ingesting a 5-ASA drug. However this methodology cannot be considered to definitively identify risk factors. Rather, it acts as a filter for highlighting the most interesting.

Potential areas of future work are developing a way to tune the minimum support used to identify the frequent itemsets and applying the methodology to a range of known prescription side effects to determine its robustness.

\bibliographystyle{IEEEtran}
\bibliography{refs}

\end{document}